\documentclass[11pt]{article}

\usepackage[preprint]{acl}

\usepackage{times}
\usepackage{latexsym}
\usepackage{framed}

\usepackage[T1]{fontenc}

\usepackage[utf8]{inputenc}

\usepackage{microtype}

\usepackage{inconsolata}

\usepackage{graphicx}
\usepackage{graphicx}
\usepackage{amsmath}
\usepackage{adjustbox}
\usepackage{amssymb}
\usepackage{booktabs}
\usepackage{algorithm}
\usepackage{algpseudocode} 
\usepackage{multirow} 

\usepackage[capitalize]{cleveref}

\title{MedCLM: Learning to Localize and Reason via a CoT-Curriculum in Medical Vision-Language Models
}

\author{Soo Yong Kim\textsuperscript{1,5,$\dagger$}, Suin Cho\textsuperscript{2,5,$\dagger$}\thanks{Corresponding Author: scho1@bu.edu}, Vincent-Daniel Yun\textsuperscript{3,5,$\dagger$}, Gyeongyeon Hwang\textsuperscript{4,5,$\dagger$} \\ \\
\textsuperscript{1}A.I.MATICS Inc, Seoul, South Korea \\
\textsuperscript{2}Boston University, MA, United States \\
\textsuperscript{3}University of Southern California, CA, United States \\
\textsuperscript{4}Heuron, Seoul, South Korea \\
\textsuperscript{5}MODULABS, Open Neural Networks Research Lab, Seoul, South Korea \\
\textsuperscript{$\dagger$}Equal Contribution
}

\begin{document}
\maketitle
\begin{abstract} 

Bridging clinical diagnostic reasoning with AI remains a central challenge in medical imaging. We introduce MedCLM, an automated pipeline that converts detection datasets into large-scale medical visual question answering (VQA) data with Chain-of-Thought (CoT) reasoning by linking lesion boxes to organ segmentation and structured rationales. These contextual signals enable medical vision-language models to generate question–answer pairs with step-by-step reasoning. To utilize this data effectively, we propose an Integrated CoT–Curriculum Strategy composed of an Easy stage with explicit lesion boxes for visual grounding, a Medium stage that encourages implicit localization, and a Hard stage for weakly supervised reasoning. Experimental results demonstrate that MedCLM attains state-of-the-art performance on several medical VQA benchmarks, providing a scalable framework for developing clinically aligned medical vision–language models. The GitHub repository will be released upon paper acceptance at: \url{https://github.com/anonymous/medclm}

\end{abstract}

\section{Introduction} 

Medical Vision Language Models (VLMs) are essential for clinical decision support. They enable systems that answer queries directly from medical images. Medical Visual Question Answering (VQA) is a central task in this field~\cite{vqarad,pathvqa,pmcvqa}. Early datasets such as VQA RAD~\cite{vqarad} and PathVQA~\cite{pathvqa} established the foundation but remain limited in scale and reasoning depth due to costly expert annotation. SLAKE~\cite{slake} and PMC VQA~\cite{pmcvqa} expanded coverage yet most benchmarks still focus on short question answering without explicit diagnostic reasoning. This limits interpretability and clinical trust.  

Chain of Thought (CoT) prompting~\cite{cot_wei} improves reasoning in large language models by producing intermediate steps~\cite{selfconsistency}. It has been effective across multimodal domains~\cite{llava,blip2} and is particularly relevant to medicine where reasoning aligns with clinical workflows~\cite{medpalm}. However constructing large scale CoT data remains costly due to dependence on proprietary models and few shot generation.  

We introduce \textbf{MedCLM}, a unified framework that integrates automatic data construction and curriculum based fine tuning for medical VLMs. MedCLM converts detection datasets into large scale VQA corpora enriched with clinically grounded CoT rationales. Structured metadata such as lesion type, location and organ provides factual seeds that guide VLMs to generate valid rationales~\cite{deeplesion,radgraph}. This removes the need for manual annotation and ensures scalability.  

To improve stability during training we employ an Integrated CoT Curriculum Strategy. Curriculum learning (CL)~\cite{bengio_cl} enhances convergence by presenting data from easy to hard. Our strategy follows this principle. The Easy stage uses explicit boxes for grounding. The Medium stage applies implicit localization with regularizers~\cite{wsddn,cutmix}. The Hard stage trains only on final answers under weak supervision~\cite{cam,gradcam}. This gradual supervision reduces cognitive load and promotes spatial reasoning without direct annotation.

\paragraph{Contributions}
We summarize our work in three main components: data construction, training strategy, and empirical validation. These components form a unified framework for building scalable and interpretable medical VLMs that remove the need for manual annotation and generalize across tasks.
\begin{itemize}
\item \textbf{Organ-aware VQA–CoT generation.} From detection datasets, we build a large VQA–CoT corpus by linking each lesion to its host organ, forming factual seeds, and prompting a medical VLM—no manual annotation.
\item \textbf{Integrated CoT–Curriculum with scheduling.} A three-stage recipe (Easy -> Medium -> Hard) separates grounding from reasoning; a domain-aware scheduler and implicit-localization regularizers stabilize training under weak supervision.
\item \textbf{Effectiveness \& interpretability.} The approach improves standard medical VQA benchmarks and radiology report generation, while producing concise, anatomically grounded rationales without extra labels.
\end{itemize}

\section{Related Work} 
\label{sec:background} 
Our work is situated at the intersection of medical visual question answering, Chain-of-Thought reasoning, and curriculum learning for vision-language models.

\paragraph{Medical AI.} With the rapid growth of AI in medicine, a wide range of analytical and predictive applications are now being developed to support clinical practice~\cite{cruzroa2017, Han2020, Hameed2022, yun2024}. Building on the success of ChatGPT~\cite{openai2023gpt4} and open-source instruction-tuned LLMs in the general domain, several biomedical LLM chatbots have also emerged, including ChatDoctor~\cite{chatdoctor}, Med-Alpaca~\cite{visualmedalpaca}, PMC-LLaMA~\cite{pmc-llama}, Clinical Camel~\cite{toma2023clinicalcamel}, DoctorGLM~\cite{doctorglm}, Huatuo~\cite{huatuo}, LLaVA-Med~\cite{llava-med}, and MedVP~\cite{medvp-llava}. These models are typically initialized from open-source LLMs and then fine-tuned on biomedical instruction-following datasets. As a result, they show strong potential for various medical applications, such as interpreting patients’ needs, assisting with biomedical analysis, and providing informed advice.

\paragraph{Medical VQA Datasets. } Medical VQA plays a key role in clinical decision support. Early datasets such as VQA-RAD~\cite{vqarad} and PathVQA \cite{pathvqa} provided curated image–question–answer pairs,~\cite{pmcvqa,medvint} but remain limited in scale, diversity, and reasoning depth due to costly expert annotation~\cite{selfrationalization}. SLAKE ~\cite{slake} introduced richer semantic labels but still lacks explicit diagnostic reasoning~\cite{pmcvqa,pmcclip}. We address these gaps with an automated pipeline that generates large-scale VQA datasets enriched with structured rationales, bypassing the annotation bottleneck.\cite{radgraph,medvint,llava-med}

\paragraph{Chain-of-Thought for Clinical Reasoning. } Chain-of-Thought (CoT) prompting ~\cite{cot_wei} elicits intermediate reasoning steps, improving tasks from arithmetic to symbolic reasoning~\cite{selfconsistency,least2most,scot}. Recent extensions apply CoT to VLMs, enabling multimodal step-by-step reasoning~\cite{multimodal_cot,llava,blip2,flamingo}. In the medical domain, CoT improves interpretability by mirroring how clinicians explain findings~\cite{medpalm,llava-med}. However, generating high-quality CoT data at scale remains challenging and often depends on few-shot proprietary models~\cite{medpalm,llm_instruct}. Our approach grounds CoT in structured metadata (lesion type, location, organ) to produce clinically relevant rationales at scale~\cite{deeplesion,totalsegmentator,radgraph,groundingdino}.

\begin{figure*}[t]
  \centering
  \includegraphics[width=1\linewidth]{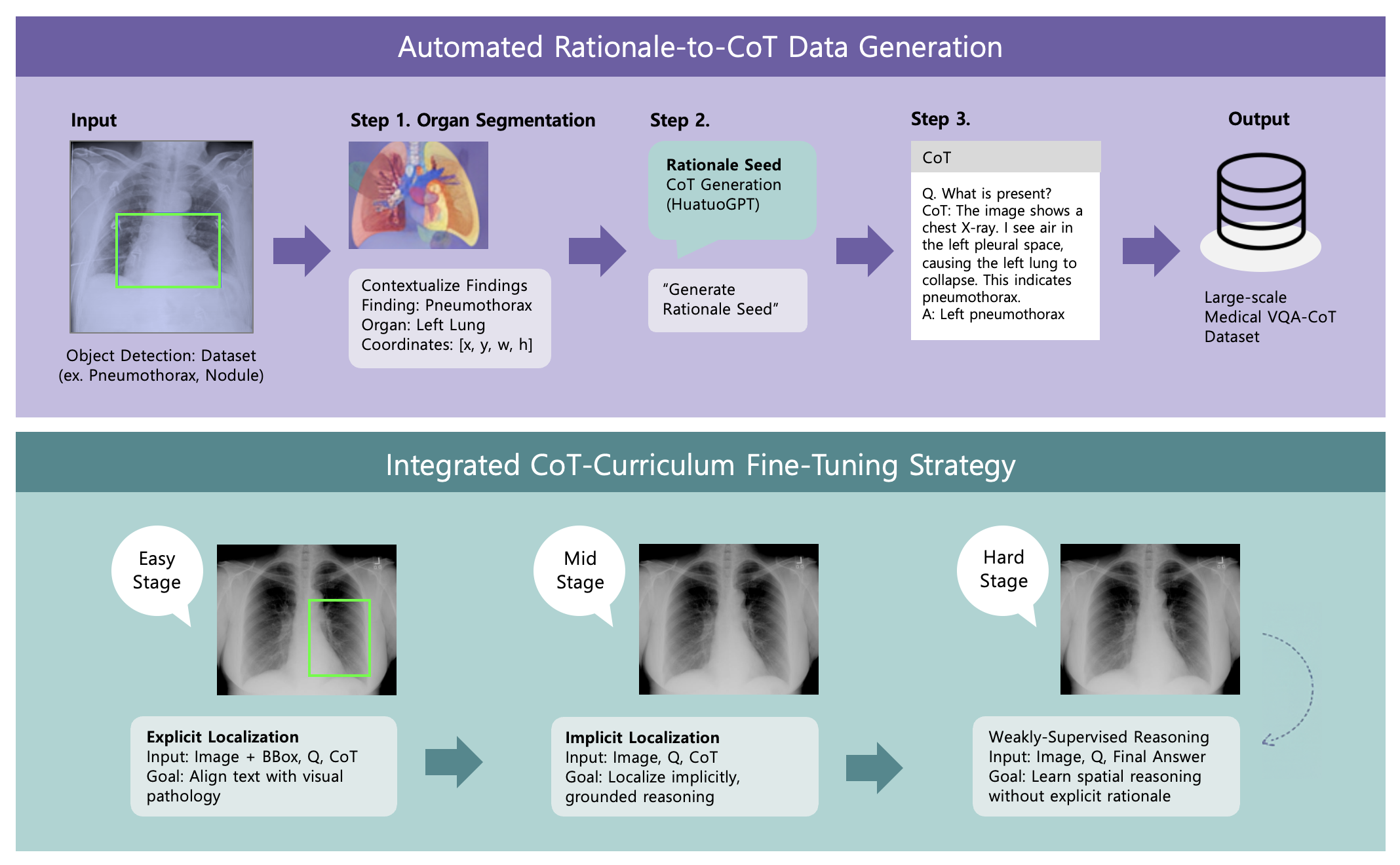}
  \caption{Automated Rationale-to-CoT Data Generation and Curriculum Fine-Tuning.
Top: Detection datasets are converted into a VQA-CoT corpus via organ segmentation, rationale seed generation, and CoT-based QA synthesis.
Bottom: Fine-tuning progresses from Explicit Localization (Easy), to Implicit Localization (Mid), and finally to Weakly-Supervised Reasoning (Hard), reducing cognitive load and improving visual grounding.}
  \label{fig:overall}
\end{figure*}

\paragraph{Curriculum Learning in VLMs. } Curriculum Learning (CL)~\cite{bengio_cl} exposes models to data in an easy-to-hard order, improving both convergence and generalization~\cite{ hacohen}. For VLMs, curricula help separate localization from reasoning, allowing models to first align visual and textual features before learning spatial grounding~\cite{clip,blip,blip2,flamingo}. Our Integrated CoT–Curriculum Strategy follows this principle~\cite{groundingdino,detr}: the Easy stage uses explicit boxes for alignment, the Medium stage enforces implicit localization with regularizers~\cite{wsddn,hideandseek,cutmix}, and the Hard stage pushes weak supervision by training only with final answers~\cite{wsddn,cam,gradcam,ovdet,attentionrollout,excitationbackprop,rightreasons}.

\section{Methodology: MedCLM}

\label{sec:method}
We present two components: (1) an automated pipeline that converts detection datasets into a CoT-enriched medical VQA corpus, and (2) an Integrated CoT–Curriculum strategy for fine-tuning VLMs. These two parts are coupled: the pipeline supplies anatomically grounded VQA–CoT data, and the curriculum schedules stage-specific objectives that progressively shift from explicit grounding to answer-only supervision.

\subsection{Automated Rationale-to-CoT Data Generation}
\label{sec:rat2cot}
\begin{algorithm*}[t]
\caption{Automated Rationale-to-CoT Data Generation}
\label{alg:pipeline}
\begin{algorithmic}[1]
\Require Detection dataset $\mathcal{D}_{\text{det}}$, organ segmentation model $\mathcal{S}$, medical VLM $\mathcal{M}_{\text{VLM}}$
\Ensure CoT-VQA dataset $\mathcal{D}_{\text{vqa}}$
\State $\mathcal{D}_{\text{vqa}} \gets \varnothing$
\For{each image $I_i$ with annotations $\mathcal{A}_i$}
  \State $\{M_k\} \gets \mathcal{S}(I_i)$ \Comment{organ masks}
  \For{each $(B_{ij}, C_{ij})\in\mathcal{A}_i$}
      \State $O_{ij} \gets \operatorname*{arg\,max}_{k}\operatorname{IoU}(B_{ij}, M_k)$
      \State $s_{ij} \gets \Call{SeedFromTriplet}{(C_{ij},O_{ij})}$
      \State $(Q_{ij}, A_{ij}, \mathrm{CoT}_{ij}) \gets \mathcal{M}_{\text{VLM}}(\Call{Prompt}{I_i,s_{ij}})$
      \State $\mathcal{D}_{\text{vqa}} \gets \mathcal{D}_{\text{vqa}} \cup \{(I_i,B_{ij},Q_{ij},A_{ij},\mathrm{CoT}_{ij})\}$
  \EndFor
\EndFor
\State \Return $\mathcal{D}_{\text{vqa}}$
\end{algorithmic}
\end{algorithm*}

\paragraph{Detection Dataset.}
We use lesion–centric corpora with \emph{bounding boxes} across CT, X-ray, and MRI. 
CT: {DeepLesion} \cite{deeplesion} (2D boxes; 32{,}735 lesions from 10{,}594 studies; +21k later annotations). 
Chest X-ray: {VinDr-CXR} \cite{vindrcxr} (18k radiographs with radiologist local labels), {RSNA Pneumonia Detection} \cite{rsnapneumonia} (pneumonia-region boxes), {NIH ChestX-ray14} \cite{nihchestxray14} (official bbox subset $\sim$984), and community {ChestX-Det} ($\sim$3.5k instance-level boxes/masks). 
Mammography: {CBIS-DDSM} \cite{cbisddsm} (updated ROIs and \emph{bounding boxes} for masses/calcifications). 
MRI: {Duke Breast Cancer MRI} \cite{dukebreastmri} (radiologist-drawn \emph{3D bounding boxes}). 
These sources satisfy the “lesion class with boxes” criterion and plug directly into our organ-aware seeding and CoT-generation pipeline.

\paragraph{Setup.}
We consider a detection dataset $\mathcal{D}_{\text{det}}=\{(I_i,\mathcal{A}_i)\}_{i=1}^{N}$, where $I_i\in\mathbb{R}^{H_i\times W_i\times C}$ is a medical image and $\mathcal{A}_i=\{(B_{ij},C_{ij})\}_{j=1}^{m_i}$ are its \emph{human-annotated} (radiologist-drawn) lesion annotations, with $B_{ij}=(x_1,y_1,x_2,y_2)\in[0,1]^4$ an axis-aligned bounding box (normalized by image size) and $C_{ij}\in\mathcal{Y}$ a lesion label. Our goal is to construct a VQA–CoT corpus $\mathcal{D}_{\text{vqa}}=\{(I_i,B_{ij},Q_{ij},A_{ij},\mathrm{CoT}_{ij})\}_{i,j}$, where $Q_{ij}$, $A_{ij}$, and $\mathrm{CoT}_{ij}$ are generated \emph{conditioned on} the lesion–organ context derived below.

\paragraph{Anatomical contextualization.}
A pretrained organ/structure segmentor $\mathcal{S}$ (we use TotalSegmentator~\cite{totalsegmentator}, CXAS~\cite{xrayanatomyseg}) produces masks $\{M_k\}_{k=1}^{K}$ for each image $I_i$. For each \emph{human} lesion box $B_{ij}$, the host organ is assigned by
\[
O_{ij}=\operatorname*{arg\,max}_{k\in\{1,\dots,K\}}\operatorname{IoU}(B_{ij},M_k),
\]
yielding the triplet $(C_{ij},B_{ij},O_{ij})$ that couples each finding with explicit organ context.

\paragraph{Seed rationale \& CoT-VQA generation.}
From $(C_{ij},B_{ij},O_{ij})$ we form a factual seed sentence $s_{ij}$ (e.g., “There is a $C_{ij}$ in the $O_{ij}$.”). Given $I_i$ and $s_{ij}$, a medical VLM $\mathcal{M}_{\text{VLM}}$ \cite{huatuo} produces a localized question, a consistent answer, and a brief rationale:
\[
(Q_{ij},A_{ij},\mathrm{CoT}_{ij})=\mathcal{M}_{\text{VLM}}\!\big(\operatorname{Prompt}(I_i,s_{ij})\big),
\]
thereby grounding CoT in the \emph{human} lesion box and the \emph{automatically selected} host organ.

\subsection{Integrated CoT--Curriculum Strategy}
\label{sec:cot-curriculum}
The curriculum stages supervision—explicit localization $\rightarrow$ implicit localization $\rightarrow$ answer-only \cite{bengio_cl, hacohen}. Let $g$ and $h$ be the visual and text encoders. Given image $I$, box $B$, and question $Q$, the model outputs a rationale $\mathrm{CoT}$ and answer $A$. We define $I'=\operatorname{draw\_box}(I,B)$, $r_B=\operatorname{ROIAlign}(g(I),B)$, and $t_{\ell,o}=h(\text{``[lesion=$\ell$] in [organ=$o$]''})$ as the lesion–organ anchor.

\paragraph{Objectives.}
We use stage-specific losses driven by \emph{training} signals. Here, $\mathcal{L}_{\text{ans}}$ is answer likelihood; $\mathcal{L}_{\text{cot}}$ is rationale likelihood (teacher-forced when provided); $\mathcal{L}_{\text{ground}}$ aligns $r_B$ with $t_{\ell,o}$; and $\mathcal{L}_{\text{attn-mask}}$ encourages model attention to overlap soft masks derived from $B$.

\paragraph{Easy (explicit localization).}
Training images include overlays $I'$, and rationales are teacher-forced. The objective combines (1) answer likelihood, (2) rationale likelihood, and (3) grounding of $r_B$ to $t_{\ell,o}$. Transition away from Easy is triggered when an EMA of the \emph{Easy-stage training loss} plateaus over $q$ consecutive epochs (see Alg.~\ref{alg:scheduler}).

\paragraph{Medium (implicit localization).}
Boxes are \emph{not} rendered to the model (no overlays visualized on image), but their masks remain in the supervision signal via $\mathcal{L}_{\text{attn-mask}}$ \cite{hideandseek, cutmix}. Concretely, we construct a soft mask $m_B$ from the box $B$ by Gaussian-blurring the binary box mask and downsampling it to the attention-map resolution, and add an alignment term $\mathcal{L}_{\text{attn-mask}}=\mathrm{KL}(\text{attn}\Vert m_B)$. Rationale supervision continues. Promotion toward Hard is considered once the \emph{Medium-stage training loss} stabilizes and a \emph{training-time rationale-loss gap} between Easy and Medium falls below a preset margin $\epsilon_{\text{cot}}$.

\paragraph{Hard CoT (answer-only reasoning).}
Only final answers are supervised: $\mathcal{L}_{\text{hard}}=\mathcal{L}_{\text{ans}}$ \cite{wsddn}. During training, multiple candidate rationales may be sampled and the one that maximizes $p(A\mid I,Q,\mathrm{CoT})$ can be used for selection, but rationales are not directly supervised.

\begin{algorithm*}[t]
\caption{Domain-Aware Curriculum Scheduler (per epoch $e$)}
\label{alg:scheduler}
\centering
\resizebox{0.9\linewidth}{!}{
\begin{minipage}{\linewidth}
\begin{algorithmic}[1]
\Require Domains $\{d\}$ (lesion class, modality); EMA rate $\rho$; ramp $\beta_e$; Hard budget $\lambda_H^{(e)}$; thresholds $(\gamma,\tau,\gamma_H,\epsilon_{\text{plat}},q,\epsilon_{\text{cot}},\delta_{\text{rise}})$; step sizes $(\eta_{\uparrow},\eta_{\downarrow})$; losses $\mathcal{L}_{\text{easy}},\mathcal{L}_{\text{medium}},\mathcal{L}_{\text{hard}}$
\State Initialize realized proportions $\lambda_E^{(e)}\gets0$, $\lambda_M^{(e)}\gets0$; keep $\lambda_H^{(e)}$ fixed within epoch
\For{each mini-batch}
  \State Sample $\lfloor \lambda_H^{(e)} B \rfloor$ Hard items from $\mathcal{D}_{\text{hard}}$; train with $\mathcal{L}_{\text{hard}}$
  \State Fill remaining slots from $\mathcal{D}_{\text{vqa}}$
  \For{each item $x$ with domain $d$}
    \State Use EMAs $m_{d}^{\text{easy},(e-1)}, m_{d}^{\text{med},(e-1)}$ of \emph{training} losses to compute
    \State $g_d^{(e)} \gets \frac{m_{d}^{\text{easy},(e-1)} - m_{d}^{\text{med},(e-1)}}{m_{d}^{\text{easy},(e-1)}+\epsilon}$
    \State $P_{\mathrm{med}} \gets \beta_e \cdot \sigma\!\bigl((g_d^{(e)}-\gamma)/\tau\bigr)$
    \State Assign $x$ to Medium w.p.\ $P_{\mathrm{med}}$, else to Easy
    \State Train with $\mathcal{L}_{\text{medium}}$ or $\mathcal{L}_{\text{easy}}$ accordingly
  \EndFor
\EndFor
\State Update per-domain EMAs $m_{d}^{s,(e)}$ from epoch-mean \emph{training} losses $\overline{\mathcal{L}}_{d}^{s}(e)$; update global EMA $\bar{m}^{(e)}$ and $\Delta\bar{m}^{(e)}\!\gets\!\bar{m}^{(e)}-\bar{m}^{(e-1)}$
\State Compute training-time rationale gap $\text{gap}_{\text{cot}}^{(e)}\!\gets\!\overline{\mathcal{L}}_{\text{cot}}^{\text{med}}(e)-\overline{\mathcal{L}}_{\text{cot}}^{\text{easy}}(e)$
\If{(\textbf{plateau}: $|\Delta\bar{m}^{(e')}|\le \epsilon_{\text{plat}}$ for last $q$ epochs) \textbf{and} $\operatorname{median}_d g_d^{(e)}\ge \gamma_H$ \textbf{and} $\text{gap}_{\text{cot}}^{(e)}\le \epsilon_{\text{cot}}$}
  \State $\lambda_H^{(e+1)} \leftarrow \min\!\bigl(\lambda_H^{(e)}+\eta_{\uparrow},\,\lambda_{H,\max}\bigr)$ \Comment{increase Hard only from training-loss signals}
\ElsIf{$\Delta\bar{m}^{(e)} \ge \delta_{\text{rise}}$}
  \State $\lambda_H^{(e+1)} \leftarrow (1-\eta_{\downarrow})\,\lambda_H^{(e)}$ \Comment{reduce Hard if total training loss rises}
\Else
  \State $\lambda_H^{(e+1)} \leftarrow \lambda_H^{(e)}$
\EndIf
\end{algorithmic}
\end{minipage}
}
\end{algorithm*}

\subsection{Curriculum Scheduling}
\label{sec:curriculum-scheduling}
The scheduler controls the per-epoch proportions $(\lambda_E^{(e)},\lambda_M^{(e)},\lambda_H^{(e)})$ of samples trained with $\mathcal{L}_{\text{easy}}, \mathcal{L}_{\text{medium}}, \mathcal{L}_{\text{hard}}$ in Sec.~\ref{sec:cot-curriculum}. A \emph{domain} $d$ is defined by lesion class and modality so that difficulty is adjusted within clinically coherent groups rather than globally. All transitions are \emph{training-loss--driven}.

\paragraph{Per-domain difficulty tracking.}
For domain $d$ and stage $s\in\{\text{easy},\text{med}\}$, we maintain an EMA of the \emph{training} loss:
\begin{align}
m_{d}^{s,(e)} \;=\; (1-\rho)\,m_{d}^{s,(e-1)} \;+\; \rho\cdot \overline{\mathcal{L}}_{d}^{s}(e),
\end{align}
where $\overline{\mathcal{L}}_{d}^{s}(e)$ is the epoch-mean of the stage-$s$ objective used for items from domain $d$. We also track a global EMA $\bar{m}^{(e)}$ of total training loss to detect plateaus and regressions.

\paragraph{Base ramp for Medium.}
A ramp factor $\beta_e$ governs when Medium samples appear:
\begin{align}
\beta_e \;=\;
\begin{cases}
0, & e\le 5,\\
\min\!\bigl(1,\tfrac{e-5}{\kappa}\bigr), & e>5,
\end{cases}
\quad (\kappa\approx10).
\end{align}

\paragraph{Adaptive assignment.}
Domain-specific progress adjusts the probability of assigning a sample to Medium; higher $g_d^{(e)}$ (i.e., smaller Medium loss relative to Easy) increases $P_{\mathrm{med}}$, shifting mass toward implicit localization. 
\begin{align}
g_d^{(e)} \;=\; \frac{m_{d}^{\text{easy},(e-1)} - m_{d}^{\text{med},(e-1)}}{m_{d}^{\text{easy},(e-1)}+\epsilon}
\end{align}
The Hard budget $\lambda_H^{(e)}$ is increased only when the \emph{training} loss plateaus for $q$ epochs, the median $g_d^{(e)}$ across domains exceeds $\gamma_H$, and the training-time rationale-loss gap $\text{gap}_{\text{cot}}^{(e)}$ falls below $\epsilon_{\text{cot}}$; it is reduced if the total training loss rises by at least $\delta_{\text{rise}}$.

\begin{table*}[t]
\centering
\scriptsize
\setlength{\tabcolsep}{4pt}
\renewcommand{\arraystretch}{0.95}
\caption{Main results on standard medical VQA benchmarks. We report \textbf{Recall} (\%) for open-ended and \textbf{Accuracy} (\%) for closed-ended questions. Our curriculum-based method achieves state-of-the-art performance across all datasets.}
\label{tab:main_results1}
\resizebox{0.7\textwidth}{!}{%
\begin{tabular}{l|cc|cc|cc}
\toprule
\multicolumn{1}{l|}{\textbf{Method}} & \multicolumn{2}{c|}{\textbf{VQA-RAD}} & \multicolumn{2}{c|}{\textbf{SLAKE}} & \multicolumn{2}{c}{\textbf{PMC-VQA}} \\
\cmidrule(lr){2-3}\cmidrule(lr){4-5}\cmidrule(lr){6-7}
\multicolumn{1}{c|}{} & Open & Closed & Open & Closed & Closed \\
\midrule
PMC-CLIP~\cite{pmcclip} & 52.0 & 75.4 & 72.7 & 80.0 & 37.1 \\
MedVInT-TE~\cite{medvint} & 69.3 & 84.2 & 88.2 & 87.7 & 39.2 \\
MedVInT-TD~\cite{medvint} & 73.7 & 86.8 & 84.5 & 86.3 & 40.3 \\
LLaVA-Med~\cite{llava-med} & 72.2 & 84.2 & 70.9 & 86.8 & 42.8 \\
LLaVA-Med++~\cite{llava-med} & 77.1 & 86.0 & 86.2 & 89.3 & \textbf{61.9} \\
MedVP-LLaVA~\cite{medvp-llava} & 89.3 & \textbf{97.3} & 91.6 & 92.9 & 58.3 \\
\midrule
MedCLM (Easy stage only) (Ours) & 89.0 & 95.9 & 91.1 & 91.8 & 59.3 \\
MedCLM (Easy $\rightarrow$ Medium) (Ours) & \textbf{90.4} & 97.1 & \textbf{92.2} & \textbf{93.4} & 61.2 \\
\bottomrule
\end{tabular}
}
\end{table*}

\section{Experiments}
\label{sec:experiments}
\subsection{Experimental Settings}
\begin{table*}[t]
\centering
\small 
\setlength{\tabcolsep}{10pt}
\renewcommand{\arraystretch}{1.2}
\caption{Main results on radiology report generation. Comparisons across two widely-used benchmark datasets, \textit{IU-Xray} and \textit{MIMIC-CXR}, using standard evaluation metrics (BLEU, ROUGE, and METEOR).}
\label{tab:report_generation}
\begin{tabular}{l|ccc|ccc}
\toprule
\textbf{Method} & \multicolumn{3}{c|}{\textbf{IU-Xray}} & \multicolumn{3}{c}{\textbf{MIMIC-CXR}} \\
\cmidrule(lr){2-4}\cmidrule(lr){5-7}
 & BLEU & ROUGE & METEOR & BLEU & ROUGE & METEOR \\
\midrule
PMC-CLIP~\cite{pmcclip} & 8.57 & 10.90 & 7.30 & 10.76 & 11.60 & 9.92 \\
MedVInT-TE~\cite{medvint} & 9.96 & 12.66 & 8.48 & 12.51 & 13.48 & 11.53 \\
MedVInT-TD~\cite{medvint} & 10.04 & 12.76 & 8.55 & 12.61 & 13.59 & 11.62 \\
LLaVA-Med~\cite{llava-med} & 9.64 & 12.26 & 8.21 & 12.11 & 13.05 & 11.16 \\
LLaVA-Med++~\cite{llava-med} & 10.82 & 13.76 & 9.21 & 13.59 & 14.64 & 12.52 \\
MedVP-LLaVA~\cite{medvp-llava} & 11.60 & 14.75 & 9.88 & 14.57 & 15.70 & 13.43 \\
MedCLM (Easy stage only) (Ours) & 11.54 & 14.67 & 9.82 & 14.49 & 15.62 & 13.36 \\
MedCLM (Easy $\rightarrow$ Medium) (Ours) & \textbf{11.73} & \textbf{14.92} & \textbf{9.99} & \textbf{14.74} & \textbf{15.88} & \textbf{13.58} \\
\bottomrule
\end{tabular}
\end{table*}
\paragraph{Datasets.}

We construct the \textsc{CoT-VQA} dataset using diverse detection datasets, leveraging its diverse lesion annotations across CT, MRI, X-Ray images. For anatomical contextualization, we employ organ segmentation models~\cite{totalsegmentator,xrayanatomyseg}. VQA performance is evaluated on three standard benchmarks: VQA-RAD~\cite{vqarad}, PMC-VQA, and SLAKE~\cite{slake}, covering different modalities (radiology, pathology) and both open- and closed-ended questions. We also evaluate the report generation performance on IU-Xray~\cite{iuxray} and MIMIC-CXR~\cite{mimiccxr} to assess report quality-factual consistency and clinical completeness. We include both VQA and report generation datasets as they assess complementary aspects of clinical image understanding: VQA benchmarks provide short-form supervision with explicit correctness criteria; report-generation corpora provide document-style supervision that emphasizes discourse coherence. Using both yields a balanced evaluation across structured QA and narrative reporting settings.


\paragraph{Implementation}
We build on \textbf{VIP-LLaVA}\cite{vip-llava} which is 7B parameters and train with AdamW under a cosine-annealing schedule with linear warm-up (initial LR $2\times10^{-5}$, $\eta_{\min}=10^{-6}$, warm-up ratio $3\%$), weight decay $0.05$, and $(\beta_1,\beta_2)=(0.9,0.98)$. We apply gradient clipping at $1.0$ and mixed precision (bfloat16 when supported, otherwise fp16); the batch size is $1$ per GPU. For our curriculum scheduler, we set the plateau patience to $q=5$ epochs and the rationale-loss gap margin to $\epsilon_{\text{cot}}=0.05$. Training begins with an Easy-only warm-up for $\sim5$ epochs, after which harder samples are gradually introduced.

\subsection{Main results}
\paragraph{Medical VQA.}
Our Integrated CoT–Curriculum achieves strong and consistent gains across VQA-RAD, SLAKE, and PMC-VQA (Table~\ref{tab:main_results1}\cite{vqarad,slake,pmcvqa}). The largest improvements appear on open-ended questions, where our method sets new state-of-the-art scores on VQA-RAD (Open) and SLAKE (Open/Closed), while remaining near–state-of-the-art on VQA-RAD (Closed) and competitive on PMC-VQA (Closed). We attribute this to the staged design\cite{bengio_cl,hacohen}: the Easy stage secures robust visual grounding, and the Medium stage enforces reasoning without explicit location cues, mitigating vague or unsupported responses while preserving accuracy on closed-ended formats.

\paragraph{Report generation.}
As shown in Table~\ref{tab:report_generation}, the Easy$\rightarrow$Medium curriculum improves report quality on IU-Xray and MIMIC-CXR\cite{iuxray,mimiccxr} over strong baselines, with consistent gains in BLEU~\cite{bleu}, ROUGE~\cite{rouge} and METEOR~\cite{meteor}. Although the gains are numerically modest, they are robust across datasets and metrics, indicating that the curriculum strategy improves the model's ability to generate factually grounded and coherent text. Qualitative analysis further shows that the model trained with our method more reliably identifies and describes lesion locations and their likely causes, moving beyond the generation capability from the generic templates toward clinically meaningful, organ aware narratives\cite{totalsegmentator,xrayanatomyseg}.


\begin{figure*}[t]
  \centering
  \includegraphics[width=0.93\linewidth]{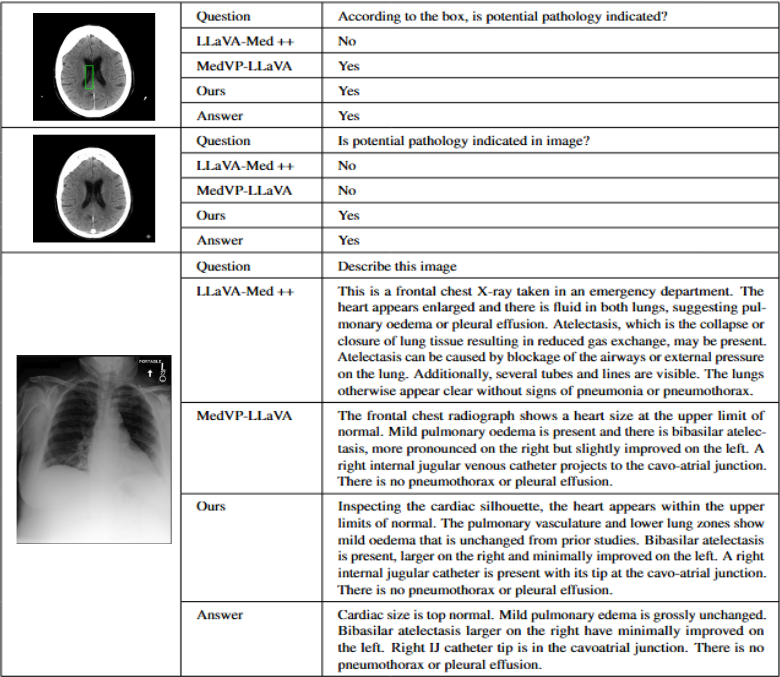}
  \caption{Qualitative comparison of model outputs on binary and descriptive medical VQA tasks. 
The first two rows show binary QA cases with and without explicit box references, where our method correctly identifies pathology while baselines fail in at least one instance. 
The third row shows a free-form description task on a chest X-ray: our model produces a clinically faithful report aligned with the reference, whereas LLaVA-Med++ introduces extraneous findings and MedVP-LLaVA omits key stability details.}
  \label{fig:qual}
\end{figure*}

\subsection{Ablation study}

\paragraph{Anatomical Rationale in Data.}
Integrating anatomical context into the data generation pipeline—by explicitly linking each lesion to its host organ—proved to be a crucial factor in improving model performance\cite{totalsegmentator,radgraph}. This contextual enrichment delivered uniform benefits across all datasets and training stages. As shown in Table~\ref{tab:ablation1}, the most significant gains were observed in open-ended question-answering on VQA-RAD and SLAKE, where the added anatomical grounding helps the model formulate more precise and relevant responses \cite{vqarad,slake}. We observed that this approach effectively reduces errors arising from anatomical confusion, such as misattributing a finding in the lungs to the mediastinum \cite{radgraph,xrayanatomyseg,totalsegmentator}.

\begin{table}[H]
\raggedleft
\caption{Ablation study on the effect of incorporating Anatomical Rationales. 
Performance comparisons on three benchmark datasets (\textit{VQA-RAD}, \textit{SLAKE}, and \textit{PMC-VQA}), 
reporting results on both open and closed-ended questions where applicable. 
AC denotes \textit{Anatomical Context} as defined in prior work~\cite{vqarad,totalsegmentator,xrayanatomyseg}.}
\begin{adjustbox}{width=1\columnwidth}
\label{tab:ablation1}
\begin{tabular}{l|cc|cc|c}
\toprule
\multirow{2}{*}{\textbf{Method}} & \multicolumn{2}{c|}{\textbf{VQA-RAD}} & \multicolumn{2}{c|}{\textbf{SLAKE}} & \textbf{PMC-VQA} \\
\cmidrule(lr){2-3}\cmidrule(lr){4-5}\cmidrule(lr){6-6}
& Open & Closed & Open & Closed & Closed \\
\midrule
Easy stage only (w/o AC) & 86.9 & 95.0 & 88.8 & 90.7 & 58.1 \\
Easy stage only (w/ AC) & 89.0 & 95.9 & 91.1 & 91.8 & 59.3 \\
Easy $\rightarrow$ Medium (w/o AC) & 88.6 & 96.3 & 90.7 & 92.5 & 60.0 \\
Easy $\rightarrow$ Medium (w/ AC) & \textbf{90.4} & \textbf{97.1} & \textbf{92.2} & \textbf{93.4} & \textbf{61.2} \\
\bottomrule
\end{tabular}
\end{adjustbox}
\end{table}

By providing organ-aware seeds, we successfully constrain the model's explanation-generation process, steering it toward clinically plausible rationales without overfitting to the specific geometry of segmentation masks \cite{rightreasons}.

\paragraph{Effect of Hard COT.}
The introduction of the weakly supervised Hard CoT stage, which relies solely on final answer supervision, yielded mixed results \cite{cot_wei,selfconsistency,multimodal_cot}. On the SLAKE dataset, this stage acted as an effective regularizer, leading to minor improvements in performance by encouraging more concise and focused rationales \cite{slake} as shown in Table~\ref{tab:ablation2}. However, on the VQA-RAD and PMC-VQA benchmarks, we observed a slight decline in accuracy \cite{vqarad,pmcvqa}. This suggests that while the Hard stage can refine reasoning when visual grounding is already robust, it may compromise answer calibration in scenarios with larger domain shifts or stronger textual priors \cite{pmcclip}. Given these findings, we adopted the more stable and consistently high-performing Easy-to-Medium curriculum for our main results, demonstrating its reliability across diverse medical VQA challenges \cite{bengio_cl,hacohen}.

\begin{table}[H]
\raggedleft
\caption{Ablation study on the effect of introducing the Hard CoT stage. Model performances with and without Hard CoT supervision across three standard benchmarks (\textit{VQA-RAD}, \textit{SLAKE}, and \textit{PMC-VQA})}

\begin{adjustbox}{width=1\columnwidth}
\label{tab:ablation2}
\begin{tabular}{l|cc|cc|c}
\toprule
\multirow{2}{*}{\textbf{Method}} & \multicolumn{2}{c|}{\textbf{VQA-RAD}} & \multicolumn{2}{c|}{\textbf{SLAKE}} & \textbf{PMC-VQA} \\
\cmidrule(lr){2-3}\cmidrule(lr){4-5}\cmidrule(lr){6-6}
& Open & Closed & Open & Closed & Closed \\
\midrule
w/o Hard COT & \textbf{90.4} & \textbf{97.1} & 92.2 & 93.4 & \textbf{61.2} \\
w/ Hard COT & 89.8 & 96.3 & \textbf{92.5} & \textbf{93.6} & 60.4 \\
\bottomrule
\end{tabular}
\end{adjustbox}
\end{table}

\subsection{Qualitative results}

Our Integrated CoT–Curriculum yields concise, anatomically consistent narratives by fostering internal spatial reasoning without overlays through staged Easy$\rightarrow$Medium$\rightarrow$Hard supervision and brief CoT steps \cite{bengio_cl,hacohen,cot_wei,selfconsistency}.

In binary QA (Fig.~\ref{fig:qual}) across \textit{VQA-RAD}, \textit{SLAKE}, and \textit{PMC-VQA}, the model correctly localizes pathology whether or not the question references a box, while baselines (LLaVA-Med++ and MedVP-LLaVA) fail in at least one case despite explicit visual prompts \cite{vqarad,slake,pmcvqa,llava-med,medvp-llava}.

For free-form description, our outputs align with key report findings (e.g., heart size at the upper limit of normal; stable mild pulmonary oedema; right-predominant bibasilar atelectasis with minimal left improvement; right IJ catheter at the cavo-atrial junction; no pneumothorax/effusion), avoiding over-calls and omissions observed in the baselines and marking progress toward clinically useful medical VLMs \cite{llava-med,medvp-llava,medpalm}.

\section{Conclusion}
\label{sec:conclusion}
We presented an automated framework that transforms detection datasets into medical VQA samples with clinically grounded Chain-of-Thought (CoT) reasoning and a structured curriculum that progresses from explicit grounding to implicit localization. This unified design encourages models to learn spatial reasoning gradually while maintaining alignment between visual evidence and textual interpretation. The framework achieves strong performance on medical VQA benchmarks, especially in open ended settings, and also improves radiology report generation by producing concise and anatomically consistent descriptions. Using a 7B backbone (ViP-LLaVA 7B), our method matches or surpasses comparable 7B models such as MedVP-LLaVA 7B and remains competitive with larger 13B variants including LLaVA-Med++. These results demonstrate that the improvements stem from the structure of the curriculum and anatomy based CoT reasoning rather than the scale of parameters.

\section{Limitations}
Our approach depends on lesion-box supervision and organ segmentation quality; errors or gaps in these inputs can propagate to CoT generation and training signals. While the Hard-stage CoT can act as a weak regularizer, its benefits are dataset-sensitive, and the most reliable default remains the Easy$\rightarrow$Medium schedule. Finally, we did not exhaustively benchmark parity-sized 13B variants or clinically validate in prospective workflows, leaving systematic size-controlled comparisons and real-world evaluation to future work.



\section{Acknowledgement}
This research was supported by Brian Impact Foundation, a non-profit organization dedicated to the advancement of science and technology for all.

\bibliography{main}


\newpage
\onecolumn
\appendix
\section{More qualitative results}
\label{sec:appendix}

\begin{figure}[h!]
    \centering
    \includegraphics[width=\textwidth, height=\textheight, keepaspectratio]{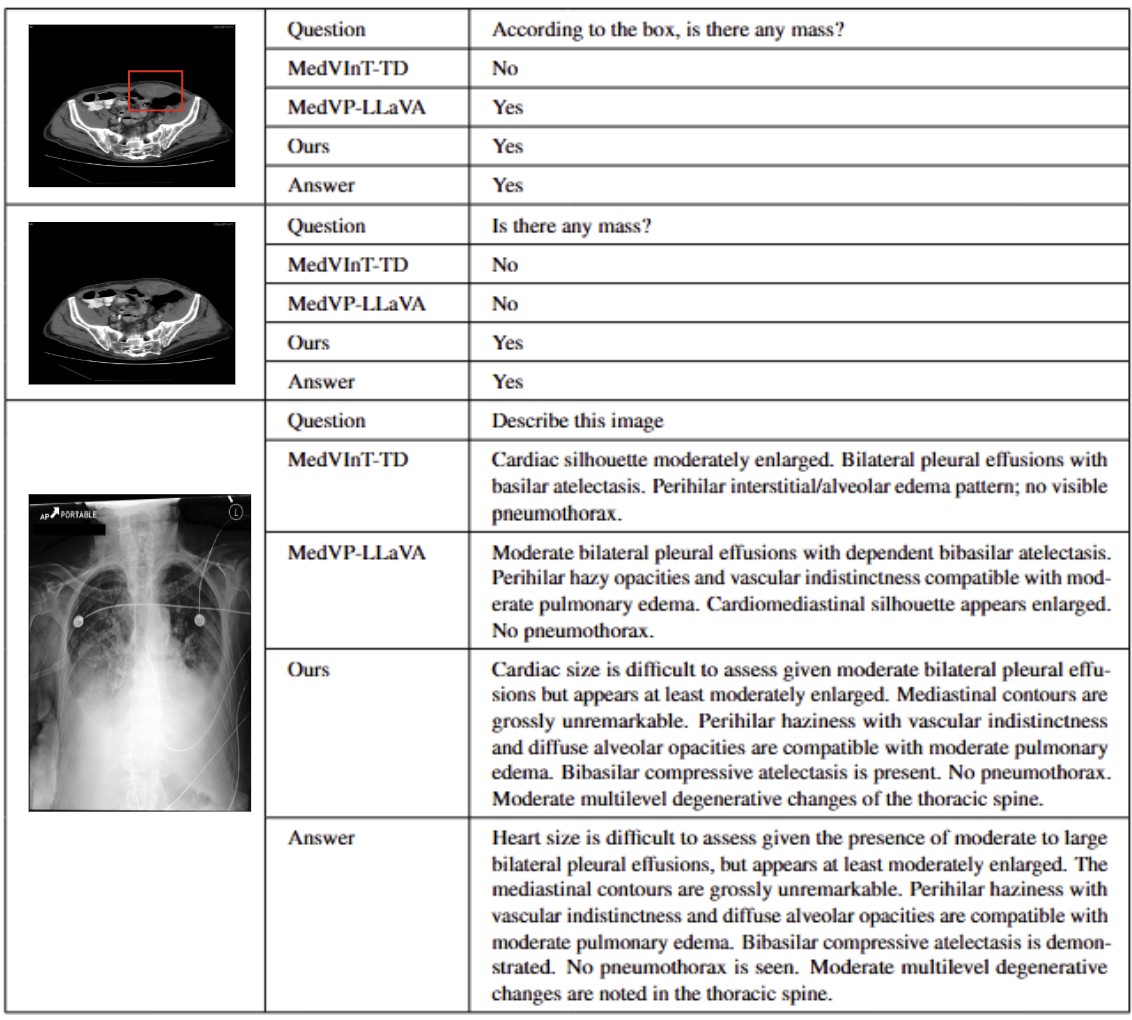}
    \caption{Additional qualitative results (1/3).}
    \label{fig:appendix1}
\end{figure}

\clearpage 

\thispagestyle{empty} 
\begin{figure}[p]
    \centering
    \includegraphics[width=\textwidth, height=\textheight, keepaspectratio]{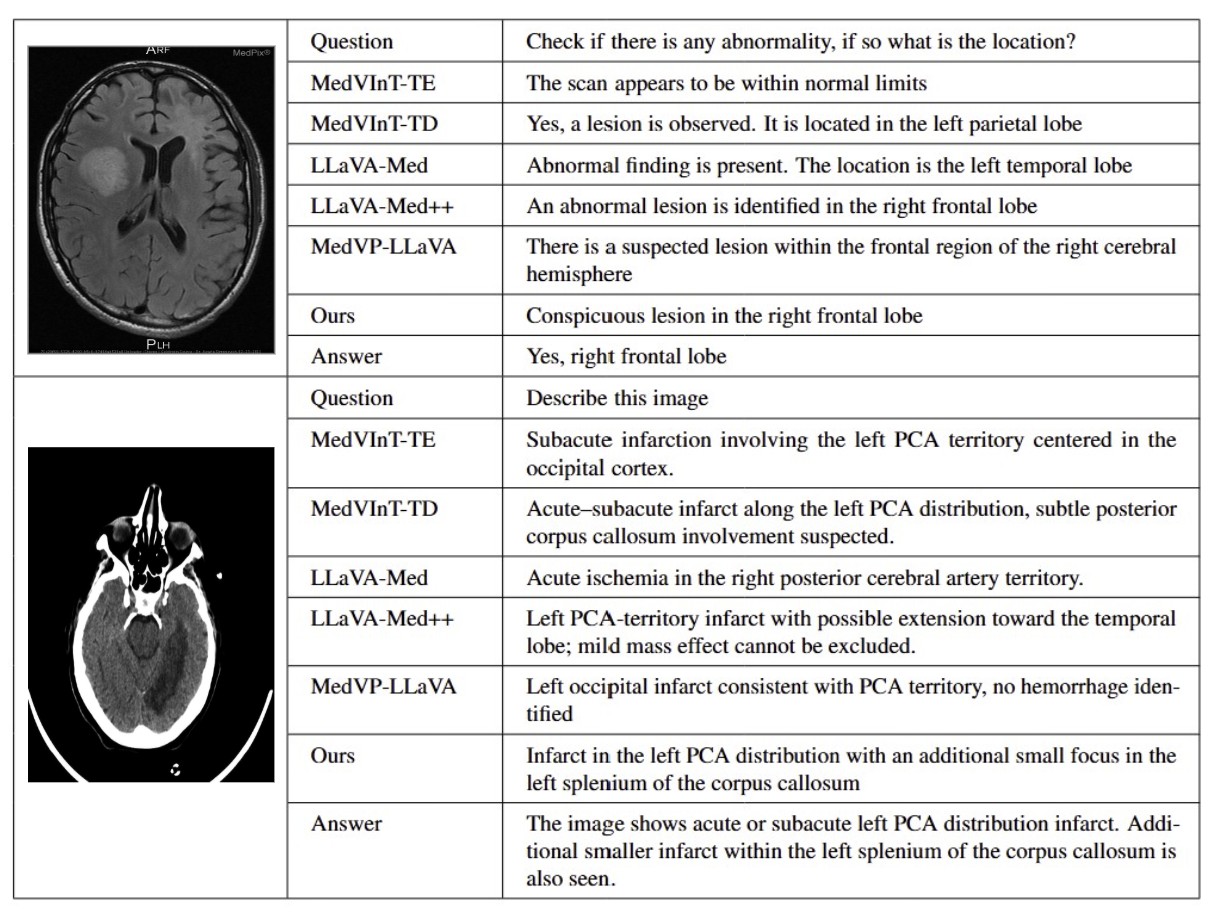}
    \caption{Additional qualitative results (2/3).}
    \label{fig:appendix2}
\end{figure}

\clearpage 

\thispagestyle{empty} 
\begin{figure}[p]
    \centering
    \includegraphics[width=\textwidth, height=\textheight, keepaspectratio]{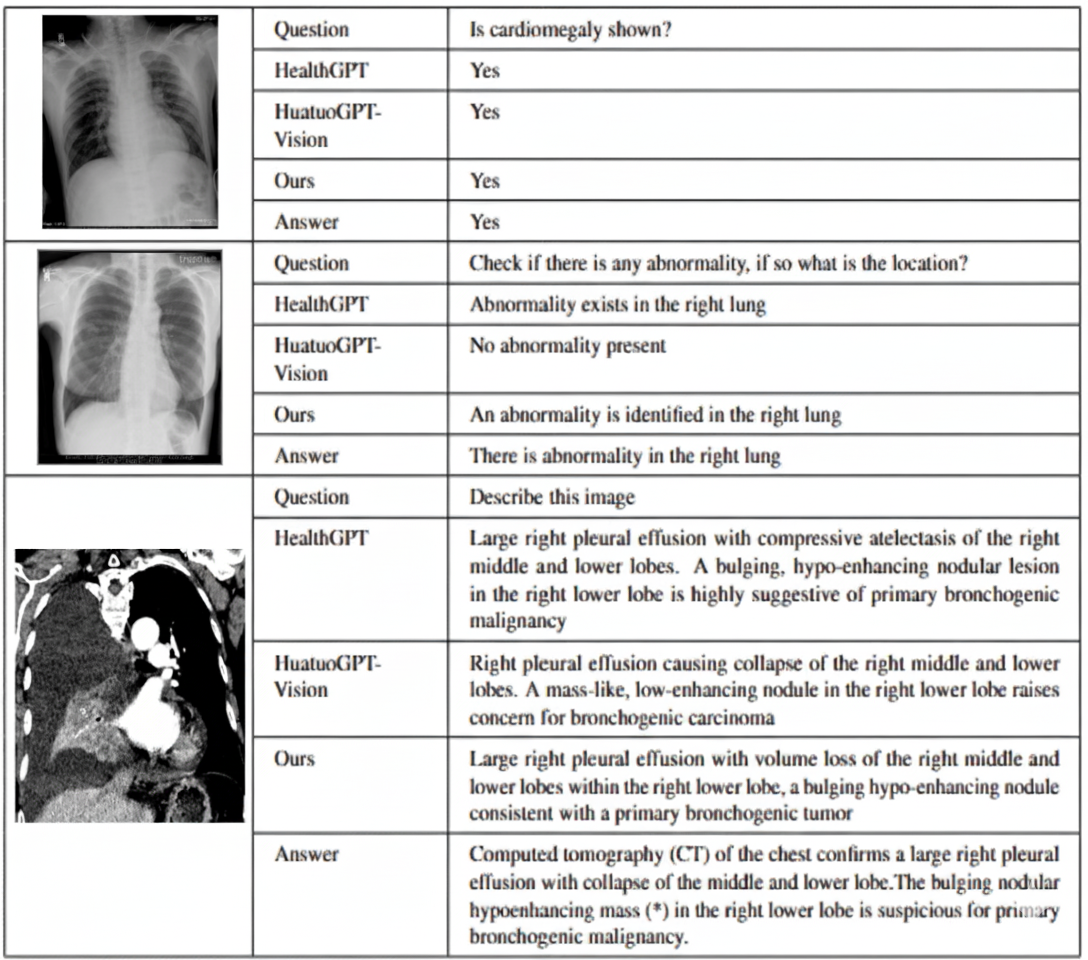}
    \caption{Additional qualitative results (3/3).}
    \label{fig:appendix3}
\end{figure}

\clearpage

\end{document}